# Can Generative Models Actually Forge Realistic Identity Documents?

*Alexander Vinogradov*
*IU International University of Applied Science*

**Abstract**-Generative image models have recently shown significant progress in image realism, leading to public concerns about their potential misuse for document forgery. This paper explores whether contemporary open-source and publicly accessible diffusion-based generative models can produce identity document forgeries that could realistically bypass human or automated verification systems.

We evaluate text-to-image and image-to-image generation pipelines using multiple publicly available generative model families, including Stable Diffusion, Qwen, Flux, Nano-Banana, and others. The findings indicate that while current generative models can simulate surface-level document aesthetics, they fail to reproduce structural and forensic authenticity. Consequently, the risk of generative identity document deepfakes achieving forensic-level authenticity may be overestimated, underscoring the value of collaboration between machine learning practitioners and document-forensics experts in realistic risk assessment.

*Disclaimer / Ethical Statement*

This study was conducted as part of the author's graduate academic work and is independent of any employer or commercial entity. No proprietary data, internal documentation, customer information, or confidential processes were used. All analyses rely exclusively on publicly available generative models and synthetic, non-identifying examples.

The purpose of this work is to examine the general capabilities and limitations of contemporary generative AI in the context of identity-document imagery. The focus is strictly analytical and defensive: understanding model behavior, identifying characteristic artifacts, and supporting realistic risk assessment for document-security research. No operational guidance, implementation details, or procedures intended to facilitate document falsification are provided.

All visual examples are based on generic, non-identifying objects and do not replicate real identity documents, protected security features, or deployed verification systems. The findings are intended solely to inform academic discussion and defensive research; any misuse would be contrary to the intent and ethical principles of this work.

This study adopts a qualitative, expert-driven forensic perspective rather than a quantitative benchmark. The selected examples are illustrative and are not intended to provide an exhaustive comparison or ranking of generative models.

## 1. Introduction

In recent years, generative AI models have triggered significant public concern regarding their potential to produce convincing fake identities and documents.

Diffusion-based architectures such as Stable Diffusion and Midjourney, and recently platforms like NanoBanano Pro and Flux can generate high-resolution and semantically consistent visual content, raising the question: can these models generate authentic-looking identity documents capable of bypassing any fraud detection systems?

This study investigates this question from both a technical and forensic perspective, combining practical experimentation with publicly available generative models and expert manual document-forensics analysis.

## 2. Background and Related Work

### 2.1 Generative AI and Information Manipulation

Generative AI can be considered a modern tool for information manipulation.

Information manipulation has a long historical context, traditionally aimed at harming or influencing societal institutions—for example, shaping public opinion or discrediting political opposition (Shoaib et al., 2023). Fake news is one of the most pervasive and impactful forms of information manipulation and can influence political discourse at scale (Allcott & Gentzkow, 2017). Historically, misinformation required substantial human effort and manual content creation to produce targeted outcomes. The advent of digital technologies began to shift this landscape, enabling broader participation in information manipulation through tools such as Photoshop, video editing software, social media platforms, and later, LM-based generative AI systems capable of rapidly disseminating such content. The combination of large language models with widely available generative models in 2025 presents a far more serious threat than what was imaginable only a few years ago. The consequences are significant and diverse: manipulated content can influence political discourse, damage personal or organizational reputations (Shoaib et al., 2023), or facilitate financial fraud (Monetary Authority of Singapore, 2025). The widespread accessibility of generative tools in 2024–2025 amplifies these risks, as virtually anyone can now create deepfake content.

### 2.2 Generative AI across modalities

Deepfakes produced by generative AI can be considered one of the core techniques of modern information manipulation. Such technology can generate a wide range of synthetic media, including speech, images, and video.

(i) Synthetic speech. It has become a practical tool for fraudsters due to advances in text-to-speech and voice-conversion technologies (Yu et al., 2023). A well-known example is the 2019 incident in which fraudsters used AI-based software to mimic a chief executive's voice and successfully request a fraudulent transfer of USD 243,000 (Stupp, 2019).

(ii) Images. Kroiß & Reschke (2025) introduced the Diverse Fake Face Dataset, demonstrating multiple face-manipulation techniques applied to address fake-news–related challenges. Deepfake image manipulation can synthesize or modify key semantic attributes, including facial identity, expression, background, or embedded text.

(iii) Video. Korshunov & Marcel (2018) presented an example of GAN-based face manipulation for video synthesis, showing how face swaps in video create significant challenges for deepfake-detection systems.

Across all of these modalities—speech, images, and video—it is increasingly difficult for an untrained observer, or even a trained one without specialized tools, to reliably distinguish between authentic and artificially generated content.

**2.3 Generative AI in FinTech and Identity Fraud**

The financial sector—particularly remote identity verification (KYC)—is one of the most affected domains. Monetary Authority of Singapore (2025) analyzes how deepfakes affect biometric authentication, targeted social engineering, and corporate digital risk and have already caused financial losses and reputational damage to organisations worldwide.

Deep fakes can:
(i) Impact Financial Institutions through market risk, cyber risk, fraud risk, regulatory risk and reputational risk;
(ii) Be used by criminals to create falsified documents, and carry out impersonation scams, enabling unauthorized transactions, financial fraud, and monetary loss;
(iii) Amplifies the potential for social engineering and impersonation scams by creating highly realistic and convincing manipulated media, such as impersonating individuals within the victim's trusted circle21. Attackers can potentially manipulate individuals into performing actions, such as transferring funds, clicking malicious links, granting system access, or divulging sensitive information;
(iv) Be used during user onboarding, to defeat verification systems by employing synthetic media and forged documents to impersonate legitimate users or create false identities. This may involve creating fraudulent documents with the use of face-swaps, face reenactments, altering facial expressions, or even creating entirely new synthetic faces.

Identity documents are now gaining increasing attention in the literature, driven by their critical role in verification workflows. In the following section, we briefly review the most impactful academic works in this emerging area and discuss their connection with generative AI.

**2.4 Academic research on identity documents**

Overall, the reviewed academic works indicate that, between 2019 and 2025, generative AI has been used in identity-document research primarily for a limited set of auxiliary tasks. These include the generation of identity photographs using models such as StyleGAN and Stable Diffusion 1.5, inpainting-based removal of variable personal data to create reusable document templates using Stable Diffusion 2.0, and the generation of synthetic metadata using large language models, including ChatGPT. In most cases, generative models are applied as supporting tools for data synthesis rather than as mechanisms for producing fully realistic document forgeries. Below, we examine these identity-document–related works in more detail.

Smart Engines is one of the pioneering companies in identity-document analysis, recognition, and verification. Among their contributions is a family of datasets, including MIDV-500, MIDV-2019, and MIDV-2020. The primary goal of these datasets is to provide a comprehensive benchmark for a wide range of identity-document analysis tasks, such as text-field recognition, document localization, semantic segmentation, document classification, and the detection of document-specific elements including portrait areas, signatures, and machine-readable zones.

The source document templates for these datasets were obtained from Wikipedia and edited to remove all variable data. Generative models—specifically StyleGAN—were used to generate synthetic faces for the portrait areas, while signatures and textual fields were also synthetically created (Bulatov et al., 2021).

Although the MIDV dataset family was not originally designed for document-forensics research, it enabled the development of several forensics-oriented extensions. One such extension is the Document Liveness Challenge dataset or DLC-2021 (Polevoy et al., 2022), which focuses on presentation attack detection tasks, including screen recapturing, unlaminated color copies, and grayscale copy attacks.

Additional forensics-related datasets built upon the MIDV family include FMIDV (Al-Ghadi et al., 2023) and SIDTD (Boned et al., 2024). SIDTD is designed to address the problem of detecting manipulated identity documents by identifying unexpected changes in document text fields, or the placement of the identity portrait. This dataset focuses primarily on composite attacks, such as crop-and-replace and inpainting-based manipulations. FMIDV concentrates on the detection of copy-move manipulation detection.

Lerouge et al. (2024) introduced the DocXPand-25k dataset, focused on visual realism and imitations of real world capturing scenarios, aiming to provide a more challenging benchmark for tasks similar to those addressed by the MIDV family. Conceptually, DocXPand-25k contains digitally manipulated images created by integrating fictitious ID designs into photographs of real identity documents using advanced transfer techniques. The authors attempted to closely mimic real-world distributions of ID document images and validated this similarity using an LPIPS-based metric, demonstrating that the dataset more closely aligns with end-user captures of real IDs compared to the MIDV family. Fictitious identity documents were designed by professional graphic designers, with AI-generated personal information, combined with a diverse collection of real-world backgrounds. The authors used Stable Diffusion v1.5 to generate identity photographs for the portrait areas.

Guan et al. (2024) addressed several key limitations of the MIDV datasets and their extensions, including limited sample sizes and insufficient manipulation of critical personal fields, which reduce their effectiveness for training models to detect realistic fraud while preserving privacy. To address these issues, the authors introduced IDNet, a large-scale, privacy-preserving dataset for identity-document fraud detection. IDNet can be considered one of the most comprehensive publicly available datasets in this domain, containing 837,060 images of distinct documents and covering a wide range of representative fraud types.

In addition to crop-and-replace and inpainting operations, IDNet introduces greater diversity in fraud patterns through manipulation techniques beyond those used in FMIDV and SIDTD, including face morphing, portrait substitution, and complex textual alterations. The authors used Stable Diffusion 2.0 to remove personal identifying information from publicly available sample identity documents, while ChatGPT-3.5-turbo was employed to generate synthetic metadata.

SynthID (Tapia et al., 2025) is a recently introduced passport dataset comprising documents from three countries, designed in accordance with ICAO requirements. The generation process includes template normalization from layered Photoshop files, structured subject metadata synthesis, biometric image selection and filtering, multimodal compositing of image layers, and reconstruction of complex visual overlays such as logos and security patterns. The resulting samples are validated using face image quality assessment software. By design, the dataset aims to address the limited realism of earlier synthetic document datasets, particularly with respect to facial portrait quality and compliance with standardized document requirements—factors that previously constrained the transferability of such datasets to real identity-document scenarios. The dataset focused on presentation attacks and includes printed and recaptured from screen images. In this study, bona fide images correspond to digitally generated samples constructed using the hybrid process described above.

While the MIDV family of datasets and their derivatives (DLC-2021, FMIDV, SIDTD), as well as more recent synthetic datasets such as IDNet, DocXPand-25k and SynthID play an important role in advancing research on automated identity-document analysis, they share a fundamental limitation: the generated document samples do not reflect the physical manufacturing technologies used in real identity documents.

The MIDV datasets and their extensions are based on templates printed using laser printers and laminated. Datasets such as SynthID and IDNet represent purely digital templates. Their bona fide samples abstract away the complex manufacturing processes that transform a digital layout into a real-world identity document - a complex physical artifact with material-dependent characteristics.

In practice, the production of real identity documents involves a much broader and more complex set of technologies that cannot be replicated through high level design imitation alone. Identity document substrates may be manufactured using different materials, including secure paper and polycarbonate and employ a wide range of different printing techniques across the document background, portrait, fixed and variable data fields. Printing methods such as offset, inkjet, dye-sublimation, and laser engraving, as well as hybrid techniques, introduce distinct, method-specific textures and visual characteristics (INTERPOL, EdisonTD).

Security elements are intentionally engineered to be difficult to counterfeit and often rely on material- and texture-specific properties. For example, iridescent inks exhibit granular and glitter-like behavior under normal lighting conditions, which cannot be realistically reproduced by digital rendering that focuses only on high-level patterns. Each of these processes produces characteristic textures and micro-patterns that differ fundamentally from purely digital imagery, introducing an additional dimension of realism beyond visual design.

This limitation makes such datasets well suited for general computer-vision tasks, but less appropriate for forensic analysis and anti-spoofing research. These observations highlight the depth and specificity of the identity-document domain, while not diminishing the pioneering contributions of the reviewed datasets.

A step toward bridging this physical gap was taken with the introduction of the MIDV-HOLO dataset (Koliaskina et al., 2023), which represents the first public dataset specifically focused on hologram-based identity-document authentication. The dataset contains 300 video samples generated from 20 different fictitious identity-document templates equipped with holographic overlays. For each template, multiple attack scenarios are provided, including copies of document templates without holograms, copies with hologram patterns manually drawn using image-editing tools, printed photographs of "original" documents containing holograms, and portrait replacement attacks. Beyond hologram detection of identity documents, MIDV-HOLO can be considered a useful benchmark for broader document-forensics research and for advancing the development of more secure identity-document verification methods.

At the same time, certain manufacturing-related aspects that are insufficiently represented in academic datasets appear to be more prominently reflected in practice by underground platforms producing fraudulent identity documents. This observation underscores the gap between current academic benchmarks and real-world threat models.

## 2.5 Underground applications

An additional perspective on the use of generative AI for identity documents comes from underground platforms. One of the most widely discussed examples is a service known as OnlyFake, which attracted public attention following an investigation by 404 Media (Cox, 2024). The platform claims to use "neural networks" to generate realistic-looking images of fake identity documents, offering them at low cost.

Media reports and online discussions describe cases in which documents allegedly produced by OnlyFake were used to successfully pass identity verification on cryptocurrency exchanges and fintech platforms such as Revolut, Coinbase, and PayPal (Bits.media, 2024). At the same time, the platform publicly discourages illegal use of its images and explicitly denies responsibility for how the generated content is applied.

Based on publicly available examples and comparisons with similar services, it remains unclear to what extent generative AI is actually used as a core component of document synthesis.

A plausible hypothesis is that platforms such as OnlyFake rely primarily on predefined document templates—potentially derived from real document references or leaked materials—rather than generating full document layouts end-to-end using generative models. In this scenario, generative AI may play a limited supporting role, for example in portrait and signature generation, background blending, or rendering adjustments, while the main document structure is produced using traditional image-editing pipelines.

A similar strategy appears to be used by another underground platform, VerifTools. Publicly available samples from VerifTools rely on a fixed set of document templates combined with a limited number of predefined backgrounds. Metadata analysis of these images reveals traces of standard image-editing software, indicating that the underlying workflow is not based on generative models. Conceptually, this approach is similar to workflow such as DocXPand-25k, where identity-document templates derived from real documents are populated with synthetic identity data and blended into a limited set of real document photographs.

Overall, both platforms appear to rely primarily on real-world document templates, into which new personal information and portraits are integrated while attempting to match the expected printing techniques and security elements. Although such outputs may appear convincing at first glance, discrepancies with genuine documents remain detectable by experts upon closer examination. Similar to the academic datasets discussed earlier, generative AI in this context appears to play a limited, secondary role rather than serving as the core mechanism for full document synthesis.

## 2.6 Summary

The reviewed literature and practical examples lead to the following key observations:

(i) Generative AI can be considered a modern and increasingly accessible tool for information manipulation, capable of producing synthetic content across multiple modalities, including speech, images, and video;
(ii) These capabilities have a direct impact on the financial sector, particularly on remote identity verification and KYC workflows, where generative techniques can be used to support impersonation and document-related fraud, for example through face manipulation or portrait substitution;
(iii) In the identity-document domain, current academic research primarily applies generative AI as a supporting tool for data synthesis, such as identity-photo generation, personal data removal, and metadata generation, rather than for end-to-end document forgery;
(iv) Existing public datasets focus largely on digital design and layout similarity, while underrepresenting the physical manufacturing processes and security features that define real identity documents;
(v) Underground platforms producing fraudulent identity documents appear to rely mainly on predefined document templates and traditional image-editing workflows, with generative AI playing a limited, auxiliary role similar to that observed in academic datasets.

In the next section, we evaluate a range of open-source and commercial generative AI platforms by testing their ability to produce document images through straightforward generation attempts. The goal is to simulate a low-barrier misuse scenario—reflecting how publicly available generative tools might be employed without model development or fine-tuning—and to assess the resulting outputs from the perspective of manual document-forensics analysis.

## 3. Methodology

### 3.1 Intuition difference between face and document deepfakes

Although face deepfakes and document deepfakes arise from similar generative principles, people intuitively perceive and evaluate them very differently. For an average observer, identifying a low-mid quality face deepfake (currently generative AI already allowed to create nearly distinguishable face deepfakes) is typically easier than identifying a low-mid quality document deepfake. This discrepancy is largely explained by everyday familiarity: individuals routinely see human faces, take photographs of faces, and interact with other people, which develops a natural intuition for facial realism. As a result, characteristic deepfake artifacts—such as unnaturally smooth skin, excessive symmetry, or inconsistent lighting—are readily noticeable even without specialized training.

In contrast, document forensics is not an intuitive skill. Distinguishing a genuine identity document from a fraudulent one requires base knowledge of manufacturing technologies, substrate materials, printing methods, optical security elements, and lawful design patterns. Most people rarely examine documents at a micro-level and therefore lack the perceptual framework needed to identify subtle discrepancies in texture, engraving, holographic behavior, or layout integrity.

The experiment section therefore should be interpreted in the context of practitioner-informed forensic inspection, rather than casual high level visual inspection, to accurately assess whether generative models can approximate the physical and technological properties of real identity documents.

### 3.2 Generation

**Objective.** The objective of this experiment was to evaluate whether text-to-image generative models can produce synthetic identity-document images that resemble authentic documents at a superficial visual level. This experiment focuses on the visual plausibility of generated document images. Controlled insertion or editing of predefined textual content was not considered at this stage and is addressed separately in subsequent experiments.

**Setup.** To guide the models, a high-level prompt framework was constructed describing the desired document type, the presence of typical design elements, and the inclusion of a portrait with characteristics expected in official identity documents. The prompt specified general attributes such as the type of document, overall orientation and background context.

Because generative models often default to expressive facial imagery, the prompt explicitly required a neutral facial expression, reflecting conventions commonly used in official identification documents. Several prompt variables were systematically varied during experimentation. The document-type parameter covered multiple categories of identity documents originating from different jurisdictions. Optional stylistic elements were included selectively to assess their influence on model behavior. To discourage stylized or low-quality outputs, a set of general suppression cues was applied, conceptually steering the models away from cartoon-like, degraded, or fantastical imagery.

To ensure a broad evaluation, a diverse set of diffusion-based generative models was tested, including multiple open-source checkpoints available via Hugging Face as well as several commercial and API-accessible models.

**Key observations.** Across all evaluated models, a consistent tendency to generate images that are unrealistic or visually inconsistent with authentic identity documents was observed. Common failure patterns included:
(i) Text generation errors, such as unreadable, distorted, or nonsensical characters that resemble pseudo-text rather than meaningful information (Fig. 1);
(ii) Poor rendering of hands and fingers when documents were generated as being held in the hand (Fig. 1), often resulting in anatomically incorrect or visually disturbing shapes, a well-known limitation of diffusion-based models;
(iii) Residual cartoon-like stylization (Fig. 1), despite explicit suppression of stylized outputs;

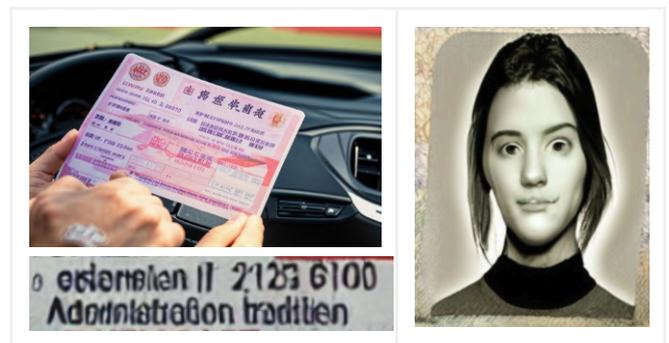

Fig. 1. Examples of common visual artifacts in generated ID documents, including anatomically incorrect hand rendering, residual cartoon-like stylization and pseudo-text.

(iv) Unnatural uniformity and lack of wear, with documents appearing overly pristine, flat, or digitally "photoshopped," lacking the subtle imperfections typical of real-world documents;
(v) Overly literal semantic associations, such as consistent insertion of bear imagery in "California" driver's license prompts (Fig. 2) or Union Jack motifs in prompts related to the United Kingdom;
(vi) Unnecessary artistic embellishments, including exaggerated reflections, dramatic shadows, lens-flare effects, or unnatural document tilting.

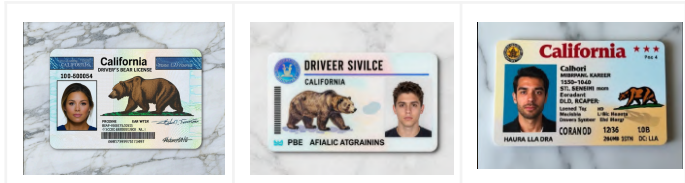

Fig. 2. Literal semantic bias in generated identity documents: multiple models associate "California" prompts with bear imagery

*From left to right: source (PII-less template), Stable-diffusion 3.5 large, Z-image turbo*

Additional examples illustrating text-to-image generation of California style driver license across different models are provided in Appendix A.

### 3.3 Manipulations

The following manipulation experiments were conducted using a limited set of visual materials selected for demonstrative purposes. These included privately owned identity documents, (which cannot be reproduced in this paper due to privacy and legal constraints), as well as a small number of physical plastic cards owned by the author (generic hologram was manually applied to approximate the visual presence of security elements) used to illustrate typical manipulation scenarios. These materials are not intended to form a dataset, but rather to provide controlled examples for analyzing model behavior across different manipulation tasks and for drawing qualitative conclusions about model limitations.

#### 3.3.1 Blending template into background

**Objective.** The objective of this experiment is to simulate a scenario in which a fraudster already possesses a clean digital template of an identity document (for example, a cropped front side of a passport or driver's license) and attempts to convert it into a plausible "live capture" by placing it into a realistic background.

**Setup.** A simple image-to-image prompt was used to instruct the models to blend the document template smoothly into a casual real-world context, such as a wooden table surface or a document held in hand, while preserving the original template without structural modification. Particular emphasis was placed on maintaining the original document layout and visual integrity, allowing only background integration and contextual rendering. Ten image-to-image model checkpoints were evaluated, including Stable Diffusion 2.x, Kandinsky 2.1, Flux-2.dev, and Nano-Banana-Pro. Representative examples of the generated outputs are shown in Fig. 3.

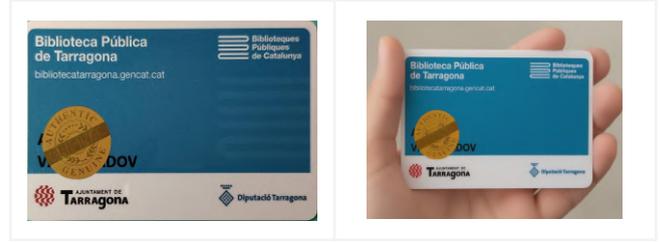

Fig. 3. Image-to-image background blending of a PII-less document template using different generative models

*From left to right: source (PII-less template), imitation of 'live' capture*

**Key observations.** The following patterns were consistently observed:
(i) Clear and systematic degradation in the fidelity of security-feature imitations. Common issues included overly smooth gradients, absence of micro-relief, and uniform shading. These artifacts resemble a digital re-drawing of the element with corresponding pattern degradation rather than a representation grounded in physical appearance (Fig. 4);
(ii) Deformation of letters (Fig. 5), introduction of non-existent symbols, and simplification or loss of fine micro-patterns;
(iii) Incorrect document scale when compared to real-world plastic card dimensions, resulting in implausible proportions and perspective inconsistencies (Fig. 6).

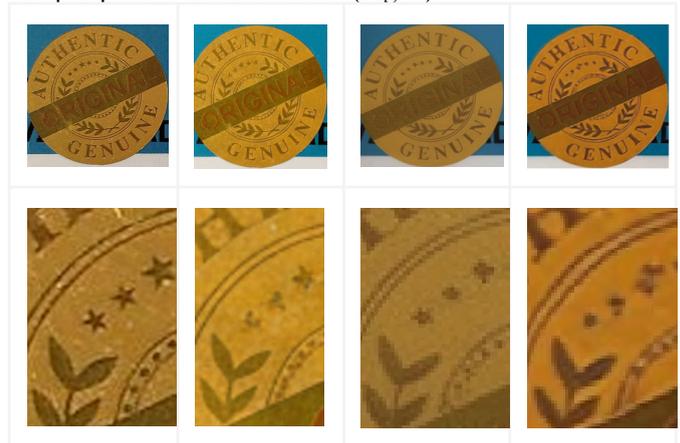

Fig. 4. Degradation of security element patterns.
*Top row (left to right): reference security element, Nano Banana, Flux-2, and Qwen-Edit. Bottom row: corresponding magnified regions highlighting texture degradation and loss of micro-relief.*

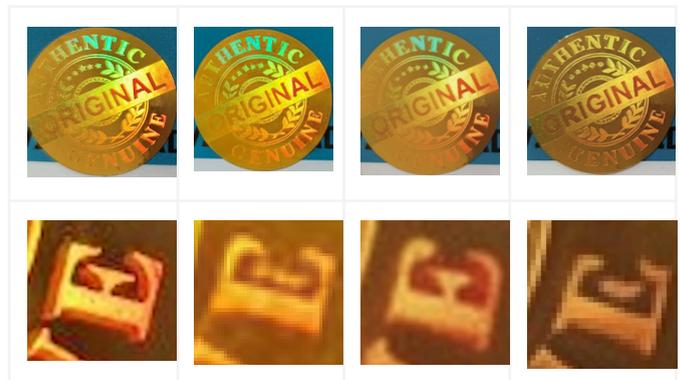

Fig. 5. Degradation of security element micro details
*Top row (left to right): reference security element, Nano Banana, Flux-2, and Qwen-Edit. Bottom row: corresponding magnified regions highlighting letter "E" geometry degradation.*

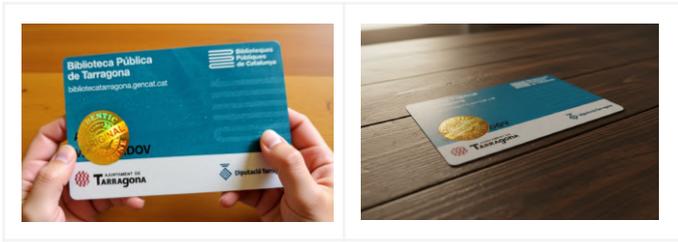

Fig. 6. Incorrect scale and implausible perspective distortion

### 3.3.2 Portrait manipulation

**Objective.** The objective of this experiment is to simulate a scenario in which a fraudster alters the portrait on the source document. The goal is not merely to paste a new portrait into the document, but to blend it smoothly into the layout and to approximate realistic portrait integration consistent with expected printing technique.

**Setup.** In the first scenario, the experiment assessed the ability of generative models to replace the portrait on document with a synthetically generated face and to approximate the visual characteristics of laser-engraving portrait integration. The models were guided to generate a portrait with a neutral facial expression in the photo placeholder and to integrate it smoothly into the surrounding document substrate (Fig. 7). The focus was on achieving an engraved-like appearance, characterized by matte texture, high local contrast, and the absence of photographic gloss or color information.

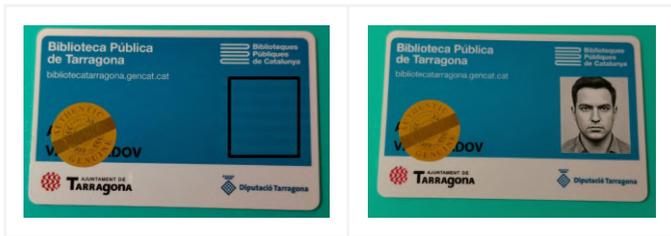

Fig. 7. Portrait manipulation using a generated portrait

*Left: source document with empty portrait placeholder. Right: model-generated and integrated portrait.*

In the second scenario, a random synthetic portrait sourced from a public face-generation website was used as transplant material and inserted into the photo placeholder. The models were guided to blend the portrait smoothly into the document and to approximate the visual appearance of inkjet-printed technique, including subtle irregular grainy ink texture, while avoiding sharp digital edges or visually detached overlays (Fig. 8).

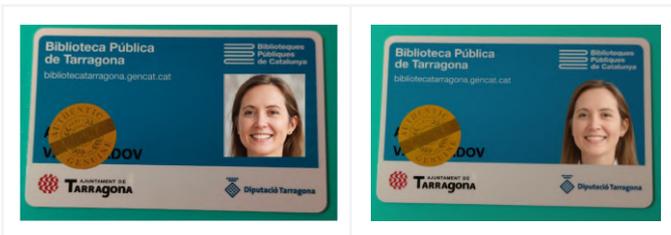

Fig. 8. Portrait manipulation using a predefined source portrait.

*Left: source document with manually pasted portrait. Right: model-refined result.*

Inpainting and image-to-image model checkpoints were evaluated, including Stable Diffusion 2.x, Kandinsky 2.1, Flux-2.dev, and Nano-Banana-Pro.

**Key observations.** Overall observations:

(i) The perceived realism of generated portraits varies substantially across models, ranging from overtly stylized or cartoon-like outputs to more photorealistic facial renderings (Fig. 9). However, increased perceptual realism does not eliminate characteristic digital artifacts characterized by excessive sharpness and overly smooth gradients;

(ii) Area modification beyond the intended manipulation region, even when region-specific masking is applied. It includes degradation of surrounding textures and text elements. In experiments involving real identity documents (examples omitted due to PII constraints), models consistently oversimplified document backgrounds by smoothing or removing existing security patterns. In particular, when security patterns extend across both the document background and the portrait area, models tend to flatten these regions into visually uniform surfaces. Fine textual elements are also affected: microtext often becomes unreadable or visibly distorted, and in some cases models introduce artificial symbols or pseudo-patterns that imitate surrounding design elements but do not correspond to any authentic document features;

(iii) the unintended modification of texture across the entire document, rather than being confined to the targeted portrait region (Fig. 10).

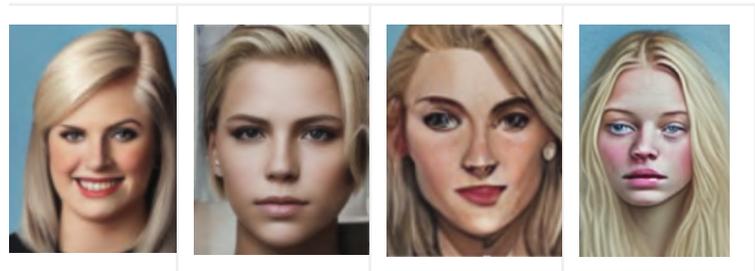

Fig. 9. Variation in perceptual realism across different generative models

*From left to right: stable-diffusion-2, realistic-vision-v4.0, dreamlike-diffusion-1.0, Kandinsky-2.1.*

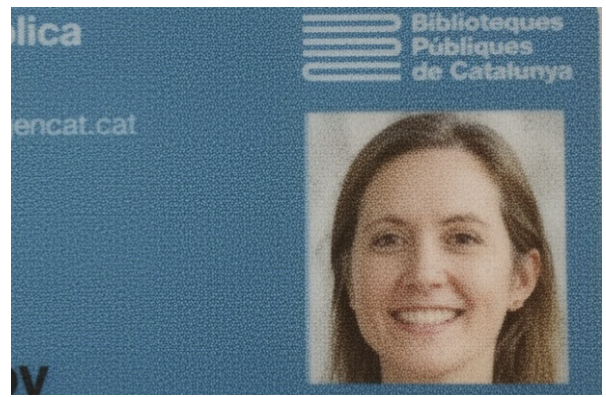

Fig. 10. Unintended global texture modification outside the portrait region

**Generated portrait substitution** (laser-engraving imitation) (Fig. 11):
(i) grayscale digital photo appearance rather than engraved texture;
(ii) artificial grain patterns that differ from laser-engraved patterns;
(iii) uniform surface texture lacking depth variation.

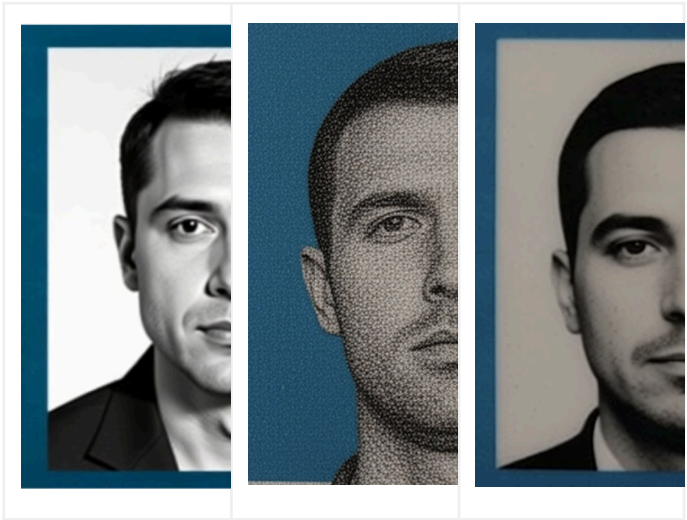

Fig. 11. Variation in perceptual realism and texture fidelity across models attempting to imitate laser-engraved portraits

*From left to right: Flux-1 context-dev, Flux-2.dev, and Qwen-Edit.*

**Predefined portrait substitution** (inkjet imitation) (Fig. 12):
(i) Distinctly digital appearance;
(ii) Textures appear algorithmically uniform, with no observable halftone or micro-dot structures;
(iii) Artificial grain uniform and regular patterns that differ from inkjet irregular patterns.

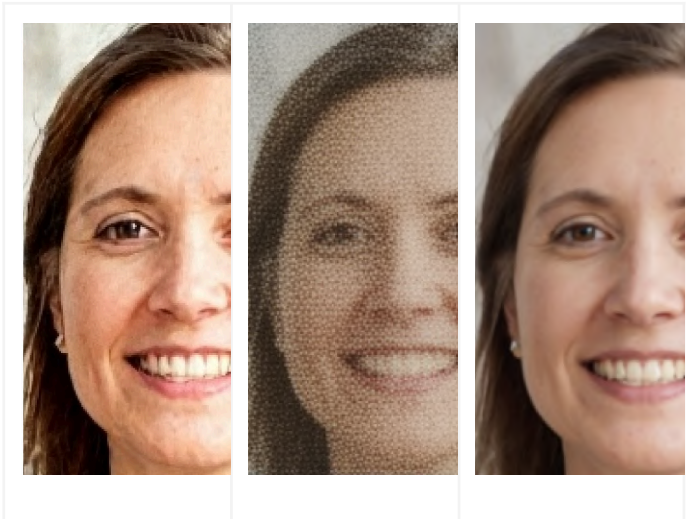

Fig. 12. Variation in perceptual realism across models attempting to imitate inkjet based portraits

*From left to right: Flux-1 context-dev, Flux-2.dev, and Qwen-Edit.*

### 3.3.3 Text manipulation

**Objective.** The objective of this experiment is to check models compatibility to reproduce pre-defined text and to paste it imitating the overall style of the provided template.

**Setup.** To generate structured text elements on the document, a high-level prompt was used to instruct the model to insert a small block of fake metadata text below an existing reference line. The prompt specified that the added text should follow the same visual style as the surrounding typographic elements—such as color, font characteristics, and relative size—and should appear as a vertically arranged set of short labels. The instruction focused on preserving layout coherence, local alignment, and stylistic consistency with the document's original design, without providing explicit operational or reproducible parameters.

**Key observations.** Overall observations:
(i) Predefined text rendering exhibits substantial instability, with quality varying significantly across models;
(ii) Visual inconsistency with the source document, including mismatches in font style, size, spacing, and color;
(iv) Unnaturally smooth, digitally rendered appearance that lacks the micro-structure characteristic of physically printed text (Fig. 13);
(v) Degradation of surrounding genuine text elements, including the hallucination of unreadable pseudo-text or malformed symbols (Fig. 14).

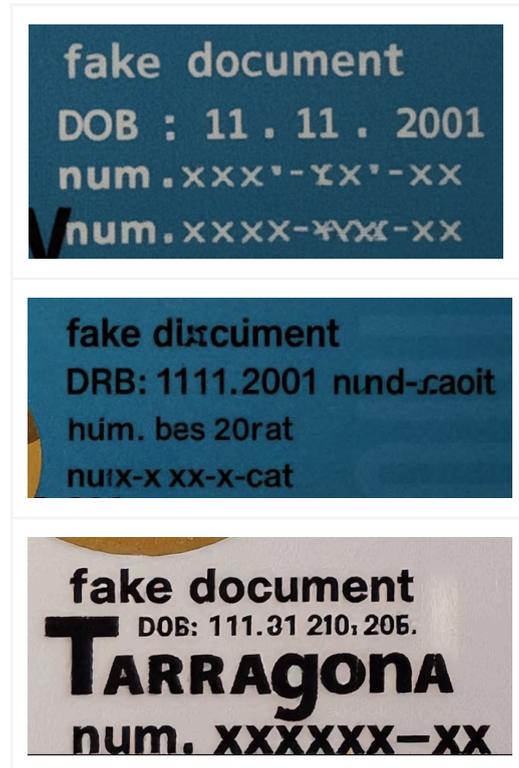

Fig. 13. Degradation and variability in predefined text rendering across generative models

*From top to bottom: , Flux-2.dev, and Qwen-edit, Kling-O1*

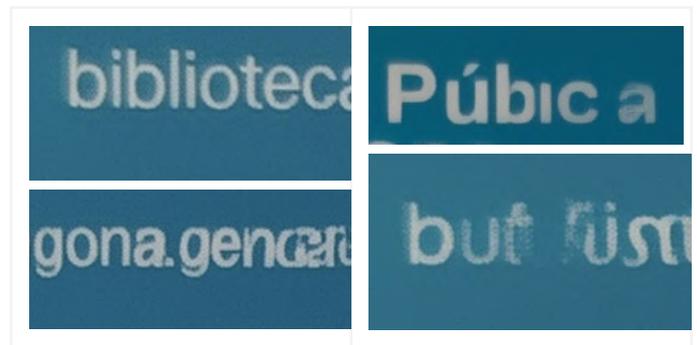

Fig. 14. Collateral degradation of non-target text regions during document manipulation

Additional examples illustrating image-to-image generation of embossed card numbers are provided in Appendix B.

### 3.4 Summary

Across all evaluated scenarios—including document generation from scratch, template integration into real-world contexts, and portrait and text manipulation—generative models exhibit a consistent pattern of behavior. When used out of the box and provided with suitable guidance or source material, modern diffusion-based models are capable of producing visually plausible document representations at a high level, reproducing overall layout, color palette, and coarse visual structure.

However, within the scope of tested scenarios these models systematically fail to reproduce the fine-grained material and manufacturing characteristics that define authentic identity documents. This limitation manifests differently across tasks: unstable and low-quality typography in text generation, digitally uniform textures in portrait substitution, and oversimplification or smoothing of security features during background integration. In all cases, the generated outputs retain a distinctly digital appearance and lack the microstructural complexity introduced by real printing, engraving, and material-processing techniques.

## 4. Conclusions and further work

This work examined the harmful applicability of modern generative models as tools for information manipulation, with a particular focus on their potential use in forging identity documents. We explored the practical capabilities and limitations of contemporary generative models from the perspective of manual document-forensics analysis.

In recent years, social media—especially LinkedIn—has become saturated with claims that modern generative models can easily produce highly realistic images of identity documents and image verification is now brittle, because generative models can recreate layouts, fonts, photos, and micro details (Thumar, 2025; Trivedi, 2025). While this trend is increasingly visible and often provocative, it also generates mixed and sometimes misleading interpretations of the actual threat level.

On one hand, as demonstrated throughout this study, generative models are indeed capable of recreating the overall visual structure of identity documents, including layout, color palette, and general design elements. On the other hand, most observers lack the expertise required to distinguish genuine documents from sophisticated forgeries. Without a detailed understanding of document-security features and manufacturing processes, visually convincing outputs may be easily mistaken for authentic documents, leading to an overestimation of the practical capabilities of generative models.

Through a series of controlled experiments, we evaluated several open-source generative models and highlighted their limitations in reproducing authentic identity documents. Our findings consistently show that generative models tend to *digitalize*, generalize, and simplify the subtle security patterns and micro-level visual features that are essential for reliable authenticity checks. As a result, while generated documents may appear convincing at first glance, they fail to replicate the physical, optical, and material characteristics that document-forensics experts rely upon during detailed examination. Consequently, in their current out-of-the-box form, generative models represent a limited threat for producing fully authentic identity-document forgeries—particularly in verification workflows that incorporate material- and texture-level auto or manual analysis. Nevertheless, this risk should not be underestimated but instead carefully assessed, mitigated, and addressed.

First, many of the fine-grained visual cues discussed in this work become significantly less distinguishable under degraded acquisition conditions, such as low resolution, blur, compression artifacts, or poor lighting. Both manual and automated document-authenticity checks depend heavily on the availability of sufficient visual information. When image quality falls below a critical threshold, reliable conclusions regarding document authenticity become increasingly difficult, regardless of whether a document is genuine or synthetically generated. As a result, low-quality captures represent a major vulnerability in verification workflows, as they may obscure precisely the cues required for robust forensic analysis.

Second, the current inability of generative models to reproduce material realism does not constitute a fundamental barrier for motivated adversaries. Existing image-processing techniques—such as style-transfer methods—can be used to partially approximate certain visual cues and conceal digital artifacts, or to hide them under low-quality capture conditions. This represents an additional risk factor that must be considered in the design of resilient verification systems.

At the same time, multiple real-world cases demonstrate successful onboarding and verification attempts using deepfake or manipulated identity documents (Monetary Authority of Singapore, 2025). These incidents suggest that some verification platforms still rely predominantly on high-level visual appearance and do not consistently perform deeper structural or security-feature analysis. This creates a critical gap: current generative technology is already sufficient to imitate the general appearance of identity documents with minimal technical expertise, while verification systems may fail to inspect the features that truly differentiate genuine documents from forgeries.

In this context, regulatory guidance—such as recommendations from the Monetary Authority of Singapore (2025)—emphasizes the need for more robust document-verification practices. Financial institutions are encouraged to verify security features such as holograms, microprinting, and optically variable inks, and to detect signs of manipulation during document validation. For example, rather than relying on a single static image, requesting a short video of the document while it is being tilted can help assess the presence and behavior of holographic elements. Additional safeguards include enforcing high-resolution image requirements, analyzing image metadata, applying image-forensic techniques to detect anomalies in lighting, shadows, or reflections, and cross-checking submitted documents against trusted data sources. Consistency checks across multiple submitted documents and scrutiny of incomplete or unverifiable customer information are also critical components of a robust verification strategy.

Future work may include the development of datasets that more accurately reflect the physical manufacturing characteristics of highly protected identity documents, while avoiding the exposure of sensitive information. Such datasets could better support the evaluation of generative-model behavior across the scenarios discussed in this work. Additionally, further research is needed to design and assess advanced verification techniques against evolving generative and post-processing attacks.

# Appendix A. Text-to-image generation of driver licenses

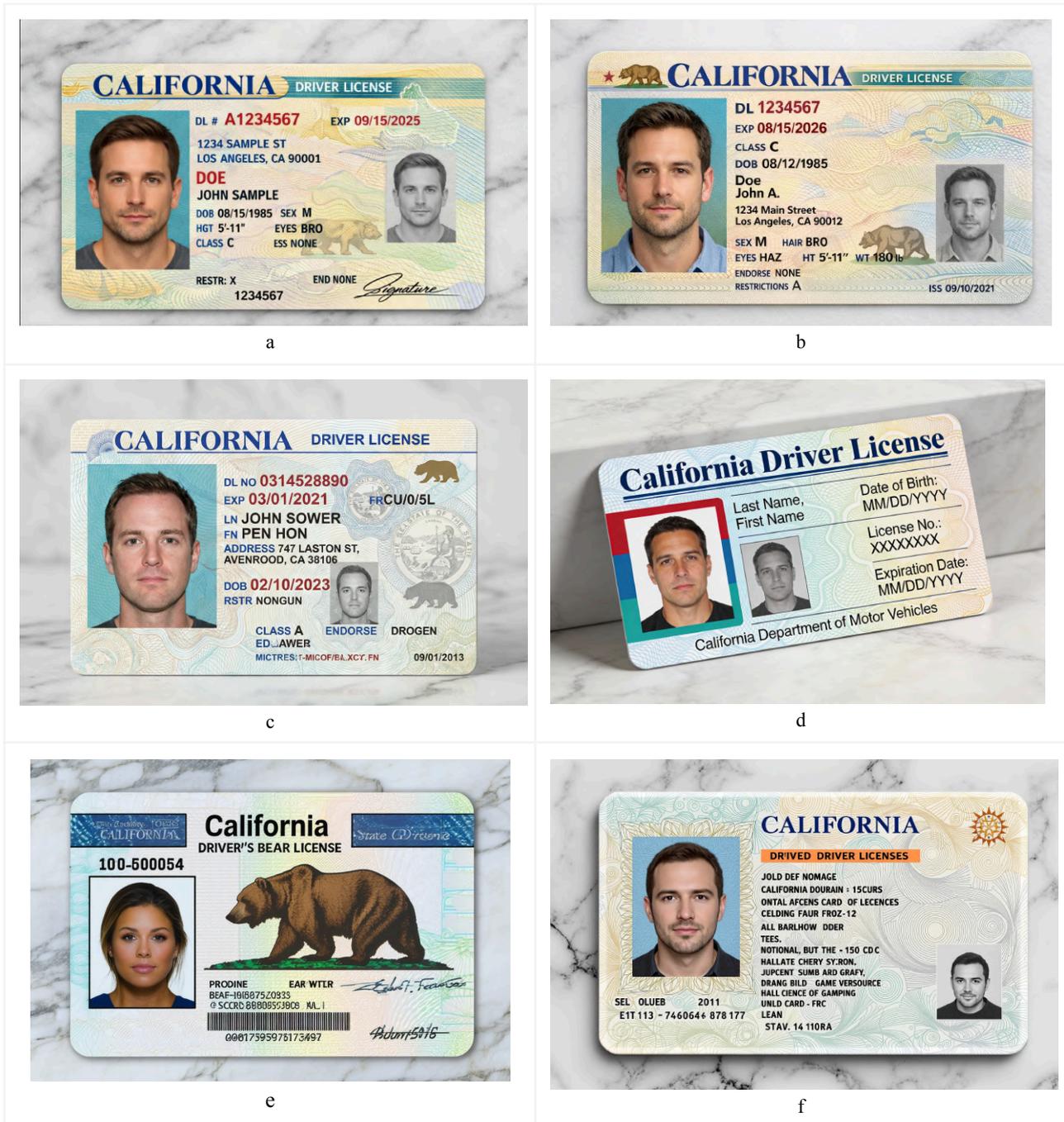

Figure A.1. Additional text-to-image generation examples for a California-style driver license across different generative models. Panels (a)–(f) correspond to representative outputs from different publicly available and API-accessed models: (GPT Image 1.5 medium, GPT Image 1.5 large, Nano Banana Pro, Seedream 4.5, Stable-diffusion 3.5, Flux 2.0 Pro)

**Appendix B. Image-to-image generation of embossed card numbers**

This appendix presents additional image-to-image generation examples illustrating text replacement and imitation of raised, embossed card numbers with physical relief. Typical failure modes include incorrect numeral geometry and font proportions (notably for digits such as *6* and *7*), overly uniform and smoothly rounded edges, and local degradation of background patterns adjacent to the modified numbers.

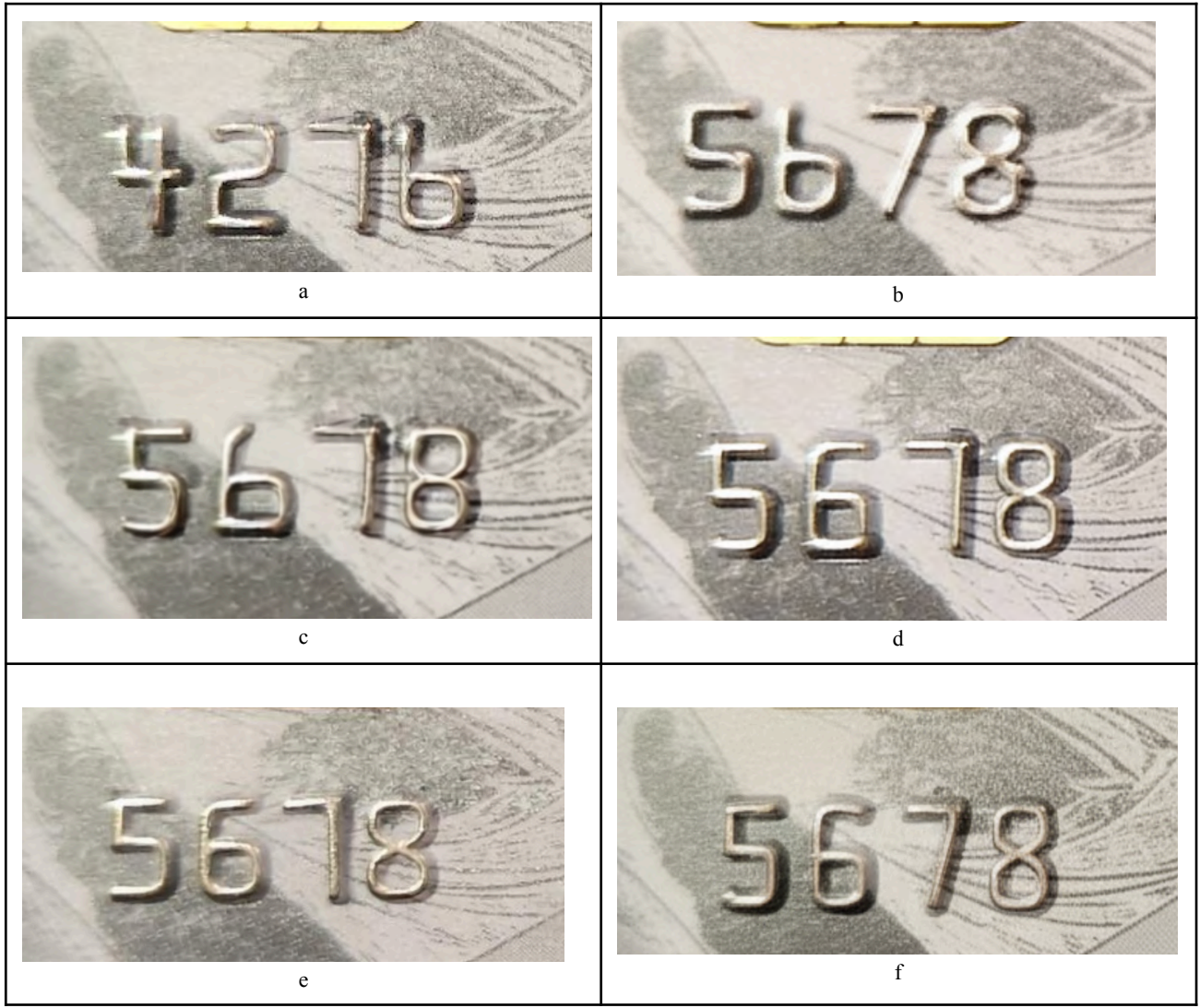

Figure B.1. Additional image-to-image generation examples illustrating replacement of embossed card numbers across different generative models. Panels (a)–(f) correspond to representative outputs from different publicly available and API-accessed models: (source, Flux-2 dev, Qwen-edit, Seedream 4.5, GPT Image 1.5 Medium, Nano Banana)